\documentclass[10pt,twocolumn,letterpaper]{article}

\usepackage{cvpr}
\usepackage{times}
\usepackage{epsfig}
\usepackage{graphicx}
\usepackage{amsmath}
\usepackage{amssymb}
\usepackage{booktabs}  
\usepackage{multirow}  
\usepackage{threeparttable}
\usepackage[symbol]{footmisc}

\usepackage[breaklinks=true,bookmarks=false]{hyperref}

\cvprfinalcopy 

\def\cvprPaperID{****} 

\begin{document}

\title{Mask Propagation Network for Video Object Segmentation}

\author{Jia Sun\footnotemark[2] ,~~Dongdong Yu\footnotemark[2] ,~~Yinghong Li,~~Changhu Wang\footnotemark[1]\\
ByteDance AI Lab, Beijing, China\\
{\tt\small \{sunjia.ring, yudongdong, liyinghong, wangchanghu\}@bytedance.com}
}
\maketitle
\renewcommand{\thefootnote}{\fnsymbol{footnote}}
\footnotetext[2]{Equal contribution.}

\begin{abstract}
   In this work, we propose a mask propagation network to treat the video segmentation problem as a concept of the guided instance segmentation. Similar to most MaskTrack based video segmentation methods, our method takes the mask probability map of previous frame and the appearance of current frame as inputs, and predicts the mask probability map for the current frame. Specifically, we adopt the Xception backbone based DeepLab v3+ model as the probability map predictor in our prediction pipeline. Besides, instead of the full image and the original mask probability, our network takes the region of interest of the instance, and the new mask probability which warped by the optical flow between the previous and current frames as the inputs. We also ensemble the modified One-Shot Video Segmentation Network to make the final predictions in order to retrieve and segment the missing instance.

\end{abstract}

\section{Introduction}

In recent years, video object segmentation has attracted much attention in the computer vision community. It has wide range of applications such as video editing, video summarization, scene understanding, and autonomous driving. Given the mask of labeled objects of the first frame, video object segmentation aims to separate the labeled objects from the background region in the future frames, which can be seen as a pixel-level object tracking task requiring fine-segmented profile and shape.

Regarding the state-of-the-art works for the task, most approaches build on basis of two mainstream methods, One-Shot Video Object Segmentation(OSVOS) and MaskTrack~\cite{caelles2017one,perazzi2017learning}. OSVOS is based on the VGG16 network which is pre-trained on ImageNet~\cite{simonyan2014very,deng2009imagenet}. At the offline stage, the network is further fine-tuned on DAVIS 2016 dataset as its parent network\cite{Perazzi2016}. Finally at the online stage, for each target video, the network is fine-tuned on the parent network by the given mask of its first frame, and used to segment the rest frames. All frames are processed independently. The results are temporally coherent and stable in the scenes where there are no drastic changes between consecutive frames. Due to the lack of temporal information, its performance decreases when it comes to some complex scenes. OSVOS shows its effectiveness in single-object segmentation, but has limitation in the multi-instance segmentation task. If there exists overlapping or occlusions among the instances, it is easy to mistake or miss all or parts of them. 

Another popular approach is MaskTrack, it predicts the mask probability of the current frame with the guidance of the mask probability of the previous frame and the image appearance information of current frame~\cite{perazzi2017learning}. In this way, it pays less attention to the useless background information and helps to separate foreground objects from background more accurately.  MaskTrack shows stability and superiority in the long consecutive videos, as temporal information propagated from frame to frame. Since MaskTrack depends on temporal continuity, changes like occlusions and pose variations are likely to degrade mask propagation process, which may lead to performance drop. If the model fails to track segmentation mask for an instance in current frame, it is difficult to recover this instance in the next frames.

\begin{figure*}
\begin{center}
\includegraphics[height = 6cm, width=18cm]{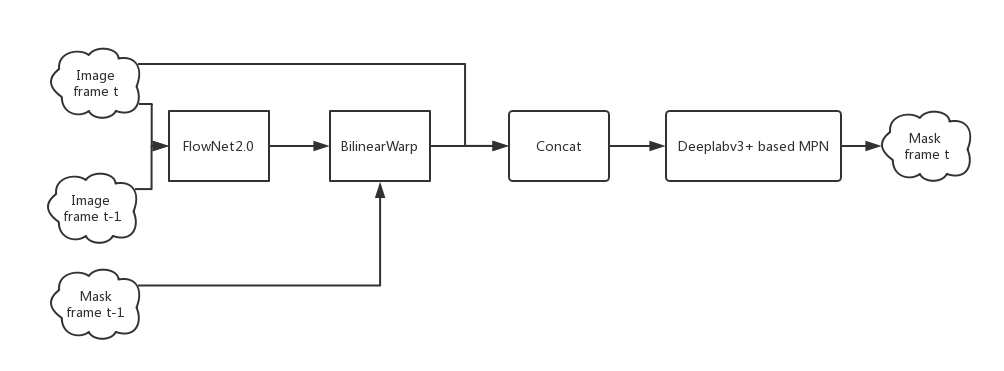}
\end{center}
\caption{Network architecture of the Deeplab v3+ based mask propagation network.}
\label{fig:1}
\end{figure*}

In order to cope with the problem, we analyze the existing approaches and ensemble the two-stream ideas to make up for each other. First, we build the mask propagation network to propagate estimated mask probability of the target object. We use the estimated mask probability of the last frame as guidance to predict the segmentation result of current frame. Second, we adopt the modified OSVOS Network to retrieve and segment the missing instance. Finally, we also apply conditional random field (CRF) on the segmentation probability map to further improve the results\cite{krahenbuhl2011efficient}.

\section{Methods}

We model the video object segmentation task as the mask propagation task based on the image appearance information and image motion information. Given two adjacent image frames $I_i$ and $I_j$, and the estimated mask probability $P_i$, we aim to predict the mask probability of the current frame $j$, $i$ and $j$ stands for two adjacent time. Inspired by the effectiveness of fully convolution network in the image classification and scene segmentation, we construct the mask propagation network to segment the identified instance by classifying each pixel into two classes: foreground instance and background. As shown in Figure \ref{fig:1}, our network is based on the state-of-the-art scene parsing network Deeplab v3+ with the following modification: the network input is replaced with the current RGB image and the previous flow guided mask probability map, the network output is specified into two classes: foreground and background ~\cite{chen2018encoder}. By using the Xception backbone based Deeplab v3+ pixel-wise classification network, we can obtain the powerful mask propagation result.

     By training the mask propagation network, given two adjacent frames $I_i$ and $I_j$, From frame $i$ to frame $j$, the estimated mask in the image $P_i$ is propagated to frame $j$, and the new mask $P_j$ is computed as a propagation function $N$ of the previous mask $P_i$, the new image $I_j$, and the optical flow $f_{i \rightarrow j}$, i.e. $P_j = N(W(f_{i \rightarrow j}, P_i), I_j)$. In this work, we first use the FlowNet 2.0 to extract the optical flow $f_{i \rightarrow j}$ from $I_i$ and $I_j$ ~\cite{ilg2017flownet}. The probability map $P_{i}$ is warpped into $P_{i \rightarrow j}$ according to $f_{i \rightarrow j}$ by a bilinear upsampling function $W$.   Then we crop the patches $P_{i,k}$, $f_{i \rightarrow j, k}$, and $I_{j,k}$ by using the bouding box of instance $k$. Rather than using the full-resolution image and flow guidance map, we feed the cropped image and flow guidance map into the mask propagation network to train the network function $N$.  This approach can leverage any existing semantic labeling architecture, such as PSPNet, ResNet, and VGG backbone based DeepLab v2 network~\cite{zhao2017pyramid,he2016deep,simonyan2014very,chen2018deeplab}. In this work, we use the state-of-the-art scene parsing network DeepLab v3+ as our mask propagation network.  
     
     We also devel a modified version of OSVOS network~\cite{caelles2017one}. We add skip-connections to every feature map before pooling operation and concatenate them after up-sampling to the output size in order to leverage multi-scale information. This allows direct supervision to feature maps of all scales, and the weights of supervision can be well controlled. To train this model, we follow the training procedure used in ~\cite{caelles2017one}. The training procedure includes two stages: the offline training stage and the online training stage at test time. At the offline stage, we first fine-tune ImageNet pretrained VGG16 network on MSRA10K dataset\cite{simonyan2014very,cheng2015global}, then fine-tune this network on DAVIS 2017 dataset to segment foreground objects from background. Since multiple instances may appear in the same frame, we simply merge instance labels into foreground label, and that improves model performance. At test time, for each instance we fine-tune the base network trained at offline stage on its label of the first frame to obtain the test network, then use this model to predict the segmentation masks of the whole video for this instance.

\renewcommand{\arraystretch}{1.8}
\begin{table*}[tp]  
  \centering  
  \fontsize{8}{8}\selectfont  
  \begin{threeparttable}  
  \caption{The performance of different methods in DAVIS 2018 Challenge.}  
  \label{tab:1}  
    \begin{tabular}{p{2cm}p{1.5cm}p{1.5cm}p{1.5cm}p{1.5cm}p{1.5cm}p{1.5cm}p{1.5cm}} 
    \toprule  
    \multirow{2}{*}{Team}&  
    \multicolumn{1}{c}{OverAll} & \multicolumn{3}{c}{ Region J}&\multicolumn{3}{c}{ Boundary F} \cr  
    \cmidrule(lr){2-2}  \cmidrule(lr){3-5} \cmidrule(lr){6-8}
    &Mean&Mean&Recall&Decay&Mean&Recall&Decay\cr
    \midrule  
    Jono&{\bf 74.7}&71.0&\textbf{79.5}&19.0&\textbf{78.4}&\textbf{86.7}&20.8\cr  
    Lixx&73.8&\textbf{71.9}&79.4&19.8&75.8&83.0&20.3\cr  
    Dawnsix&69.7&66.9&74.1&23.1&72.5&80.3&25.9\cr  
    TeamILC\_RIL&69.5&67.5&77.0&15.0&71.5&82.2&18.5\cr  
    Apata&67.8&65.1&72.5&27.7&70.6&79.8&30.2\cr  
    UIT&66.3&64.1&75.0&\textbf{11.7}&68.6&80.7&\textbf{13.5}\cr  
    Alextheengineer&60.6&58.4&65.6&26.2&62.9&71.0&29.7\cr 
    TeamVia(\textbf{Ours})&60.1&57.7&64.9&27.2&62.4&71.7&28.1\cr 
    Kthac&58.9&56.7&63.1&30.7&61.1&67.6&33.1\cr  
    \bottomrule  
    \end{tabular}  
    \end{threeparttable}  
\end{table*}

    In the inference stage, we use the threshold to generate the instance ROI bounding box. We also adopt the modified OSVOS network to retrieve and segment the missing instance. As a final stage of our pipeline, we refine the generated the mask $I_j$ using DenseCRF per frame\cite{krahenbuhl2011efficient}. This adjusts some image details that the network might not have captured.

\begin{figure*}
\begin{center}
\includegraphics[height = 20cm, width=18cm]{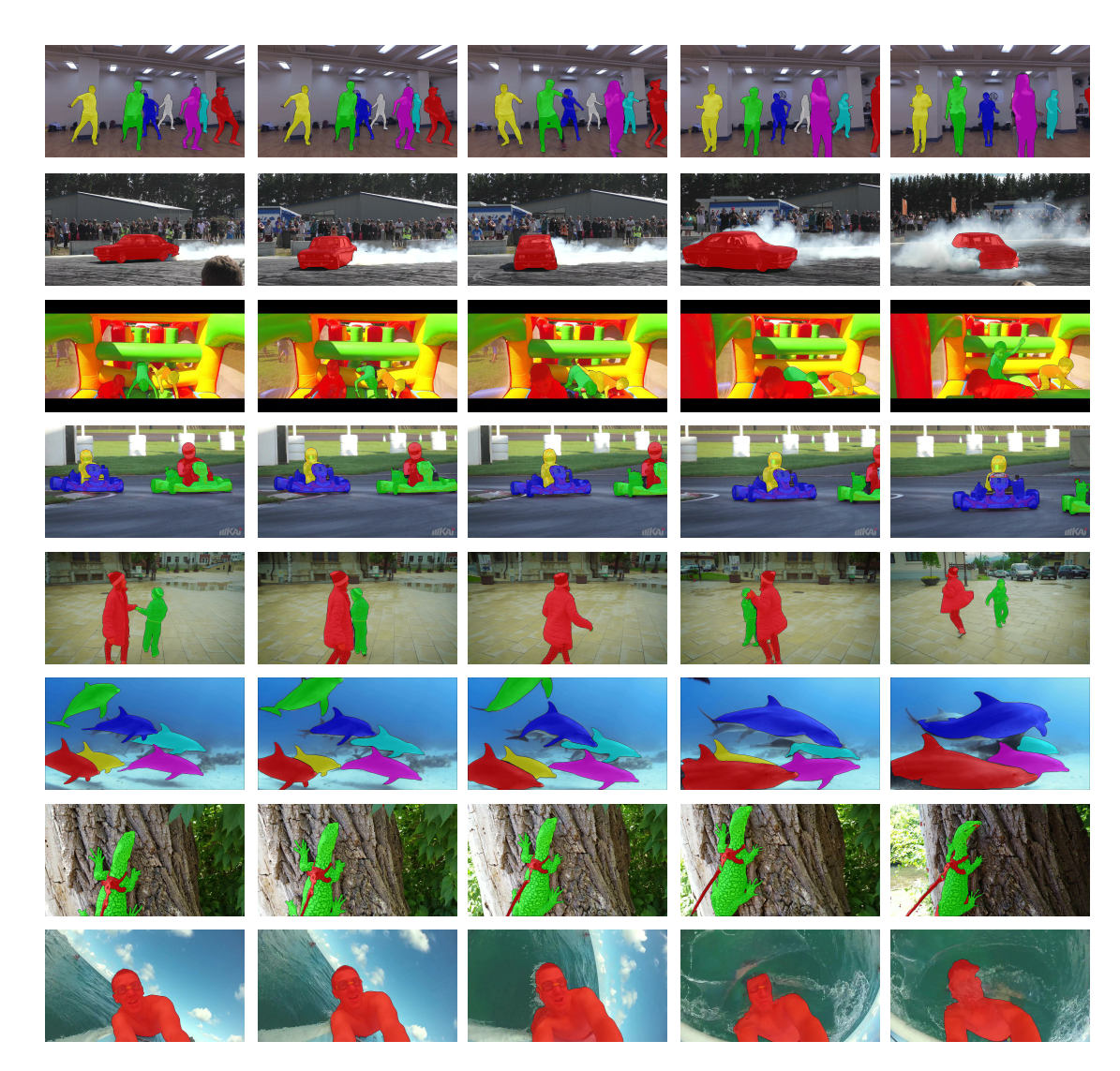}
\end{center}
\caption{Qualitative results on DAVIS 2018 test-challenge set.}
\label{fig:2}
\end{figure*}

\section{Experiments}

We evaluate our method on the DAVIS 2017 dataset, which contains 60 videos in the train set and 30 videos in the val set with pixel-level annotated object masks for all frames, and 30 videos in the test-dev set and 30 videos in the test-challenge set with the pixel-level annotated object mask for the first frame~\cite{Pont-Tuset_arXiv_2017}. In our experiments, we use training set and validation set as training data in the training process. And predictions of test-dev set and test-challenge set are submitted to the CodaLab site of the challenge for evaluation. We use Region Jaccard (J) and Boundary F measure (F) as the evaluation metrics for each instance ~\cite{Perazzi2016}. As show in Table \ref{tab:1},  we present the predicted segmentation results in the 2018 DAVIS Challenge and our score is 57.7\% in J and 62.4\% in F. The Figure \ref{fig:2}  shows some examples of our predicted segmentation results in the 2018 DAVIS Challenge.

\section{Conclusions}

In this work, we propose to use the mask propagation network for video instance segmentation. We show that on the DAVIS 2017 dataset, the proposed mask propagation network achieves competitive performance for multiple instance segmentations in videos.

{\small
\bibliographystyle{ieee}
\bibliography{egbib}
}

\end{document}